\documentclass{mva_style}
\usepackage{graphicx}
\usepackage{color}

\usepackage{amsmath}
\usepackage{amssymb}
\usepackage{subcaption}

\finalcopy %Uncomment this line for the Camera-Ready Manuscript

\begin{document}
\title{Unsupervised 3D Braided Hair Reconstruction \\from a Single-View Image\thanks{\scriptsize Accepted to the 2025 International Conference on Machine Vision Applications (MVA 2025).}
}
% \title{Unsupervised Single-View 3D Braided Hair Reconstruction}

\author{
  Jing Gao\\
  Carnegie Mellon University \\
  Pittsburgh, PA, USA\\
  {\tt jinggao2@andrew.cmu.edu}
}

\maketitle

\section*{\centering Abstract}
\textit{
  Reconstructing 3D braided hairstyles from single-view images remains a challenging task due to the intricate interwoven structure and complex topologies of braids. Existing strand-based hair reconstruction methods typically focus on loose hairstyles and often struggle to capture the fine-grained geometry of braided hair. In this paper, we propose a novel unsupervised pipeline for efficiently reconstructing 3D braided hair from single-view RGB images. Leveraging a synthetic braid model inspired by braid theory, our approach effectively captures the complex intertwined structures of braids. Extensive experiments demonstrate that our method outperforms state-of-the-art approaches, providing superior accuracy, realism, and efficiency in reconstructing 3D braided hairstyles, supporting expressive hairstyle modeling in digital humans.
  % which facilitates richer and more expressive hairstyle representations for digital human applications.
}

\section{Introduction}
\label{sec:intro}

Reconstructing diverse hairstyles is crucial in the digital human industry, as hair significantly influences an individual's appearance and self-expression \cite{chiang2015practical,piuze2011generalized,shen2023ct2hair}. Accurately capturing hairstyles, especially in 3D, enhances realism in digital avatars and animations, enabling richer personal and cultural expression. However, due to the complex geometry and intricate topology of hair, reconstructing accurate structures remains challenging, particularly for sophisticated styles such as braids.

Existing hair reconstruction methods predominantly fall into two categories: multi-view and single-view approaches. Multi-view techniques \cite{luo2012multi,nam2019strand,zhang2018modeling} achieve high-quality reconstructions but often require controlled setups and specialized hardware, limiting their practical applicability. Single-view methods \cite{hu2015single,saito20183d,shen2020deepsketchhair,wu2022neuralhdhair,zhou2018hairnet} are more practical but face significant challenges due to limited viewpoint information. Early deep learning-based single-view methods such as HairNet \cite{zhou2018hairnet}, Hair-VAE \cite{saito20183d}, and Hair-GANs \cite{zhang2018hair} heavily rely on large-scale synthetic datasets to train models capable of reconstructing general loose hairstyles. However, synthetic data often lacks sufficient realism and diversity, limiting the generalization of these methods to complex, structured hairstyles like braids. Additionally, strand-based hair reconstruction has become the standard for achieving high-quality realistic representations, which is beneficial for rendering and physics-based simulations \cite{zakharov2025human,fei2017multi,hsu2023sag,daviet2023interactive}. Recent works such as NeuralHDHair \cite{wu2022neuralhdhair}, Neural Haircut \cite{sklyarova2023neural}, and HairStep \cite{hairstep} further advance strand-level reconstruction. Nevertheless, they primarily target general hairstyles and have difficulty accurately representing complex braided structures due to insufficient structural modeling.

Braided hairstyles, despite their popularity, pose unique challenges for reconstruction due to their intricately intertwined structures. Fundamental braid theory \cite{Artin1950BraidTheory} has extensively studied the systematic braid patterns, such as fishtail, four-strand and French braids. 
Such braids follow repetitive, rule-based structures \cite{DIYbraid}, providing a systematic basis for modeling and reconstructing these intricate styles. However, practical reconstruction remains challenging due to the scarcity of 3D braided hair data, the difficulty in obtaining detailed strand-level annotations. Previous methods often rely on RGB-D inputs and database-driven synthesis \cite{hu2014capturing} to reconstruct braids. These approaches are often impractical or limited in scope, failing to generalize effectively to single-view images freely available online.
% For instance, a classic three-strand braid follows a simple yet systematic process: the left strand is crossed over the middle strand, followed by the right strand crossing over the new middle strand. Repeating this pattern continuously results in the complete braid.
% Leveraging these principles can significantly simplify modeling and reconstruction processes for complex hairstyles, enabling more diverse and realistic representations in digital environments.

To address these limitations, we propose a novel unsupervised pipeline for reconstructing strand-based braided hair directly from single-view RGB images. Our method integrates off-the-shelf models with a lightweight optimization-based braid model inspired by braid theory, using procedural sinusoidal functions to represent complex interwoven braid patterns.
% Our method integrates pretrained off-the-shelf models without additional training, focusing instead on an optimization-based braid modeling approach. Guided by braid theory, we introduce a novel synthetic braid model characterized by procedural sinusoidal functions, enabling accurate representation of complex interwoven braid patterns.
Our pipeline involves three primary stages: (1) obtaining a coarse hair estimation and a braid mask from off-the-shelf models; (2) training the synthetic braid model using learning-based optimization inspired by braid theory; (3) refining the coarse hair reconstruction based on the optimized synthetic braid model. This refinement effectively resolves structural ambiguities caused by occlusions, significantly enhancing reconstruction accuracy. 
Notably, our synthetic braid model is trained independently and efficiently, typically requiring only a few minutes per reconstruction.
Comprehensive experiments demonstrate our method's effectiveness using diverse, single-view images sourced entirely online. Our method achieves superior accuracy and realism compared to existing state-of-the-art approaches \cite{hairstep,neuralhd}, particularly excelling in reconstructing intricate braid geometries previously unattainable by single-view methods. 

The primary contributions of our work include a novel unsupervised pipeline that efficiently reconstructs complex braided hairstyles from single-view RGB images without requiring 3D ground truth, 
an innovative synthetic braid model trained via braid theory-inspired optimization for detailed braid structure representation, and extensive evaluations demonstrating superior accuracy, realism, and efficiency compared to existing methods. By addressing critical challenges in hair reconstruction, our method facilitates more diverse, accurate, and expressive hairstyle representations in digital human applications and virtual environments.

\section{Method}
% \vspace{-10pt}
\label{sec:method}

\begin{figure*}[t]
    \centering
    \begin{subfigure}[t]{0.68\linewidth}
        \centering
        \includegraphics[width=1\linewidth]{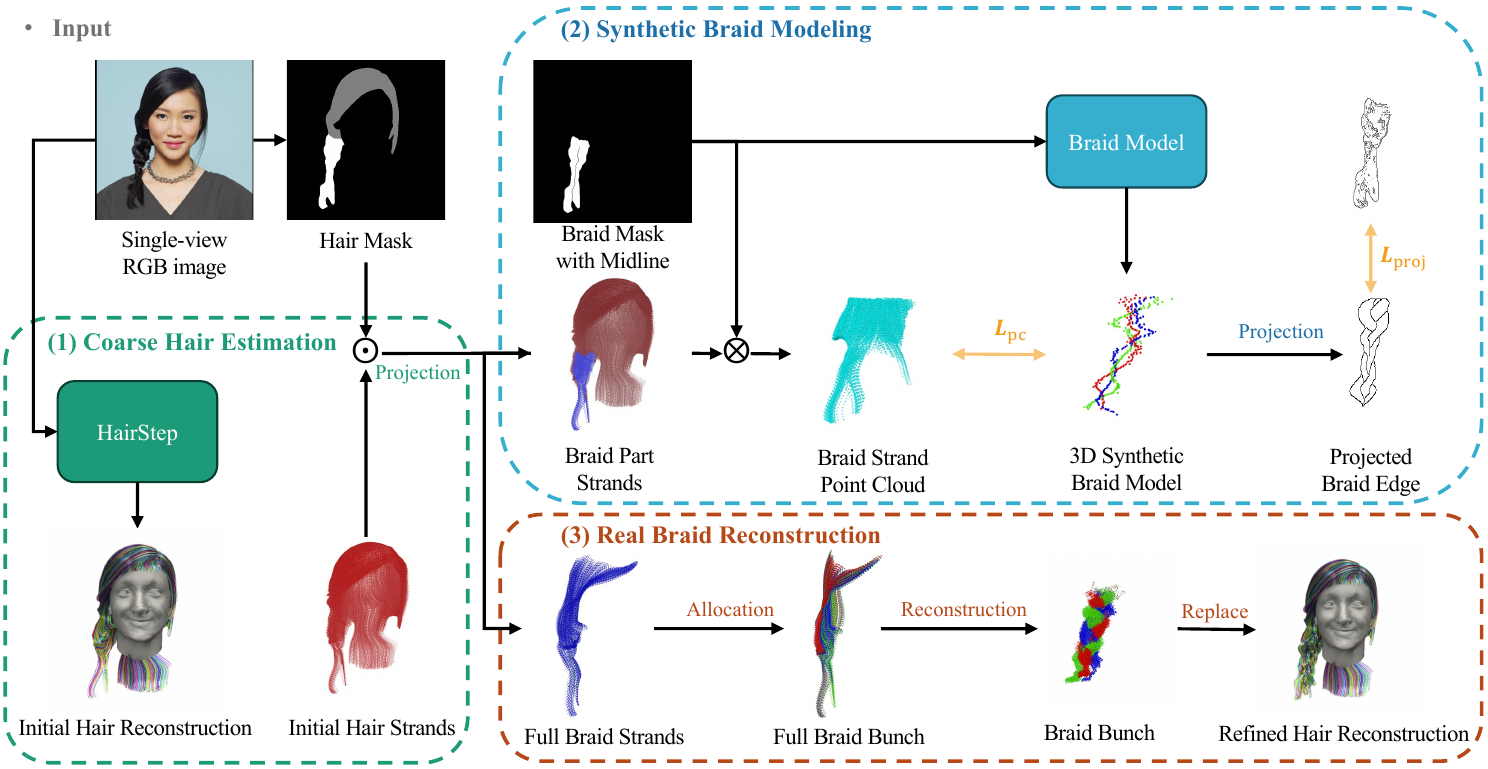}
        \small \caption{Overview of our 3D braided hair reconstruction pipeline. (1) Coarse Hair Estimation: Generates an initial hair structure with strand-based methods. (2) Synthetic Braid Modeling: Captures braid structure using a procedural model. (3) Real Braid Reconstruction: Refines strands by aligning with the synthetic model.}
        \label{fig:pipeline}
    \end{subfigure}    
    \begin{subfigure}[t]{0.173\linewidth}
        \centering
        \includegraphics[width=1\linewidth]{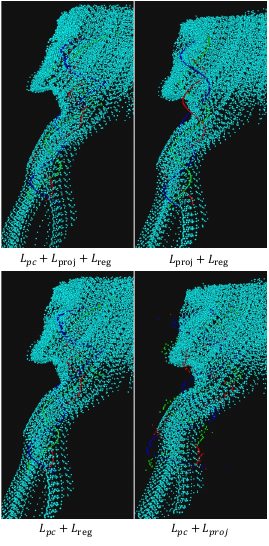}
        \small \caption{Ablation on losses.}
        \label{fig:syn+real_ablation}
    \end{subfigure}
    \begin{subfigure}[t]{0.13\linewidth}
        \centering
        \includegraphics[width=1\linewidth]{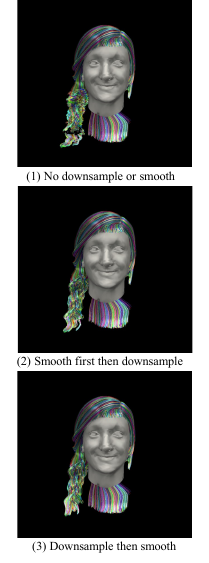}
        \small \caption{Ablation on downsampling and smoothing.}
        \label{fig:downsample_smooth_ablation}
    \end{subfigure}
    \vspace{-10pt}
\end{figure*}

Our framework introduces an unsupervised approach to capture complex braided hairstyles by refining a coarse 3D hair estimation. We represent 3D hair with strand-identifiable point clouds, where each strand is modeled as a unique set of 3D points. As illustrated in Fig.~\ref{fig:pipeline}, the pipeline consists of three phases: coarse hair estimation, synthetic braid modeling, and real braid reconstruction.
\subsection{Coarse Hair Estimation}
\label{sec:coarse}

We generate an initial 3D hair structure using a combination of strand-based reconstruction techniques, starting with the HairStep model~\cite{hairstep} to create intermediate strand and depth maps. These representations are further processed with a pre-trained occupancy network~\cite{neuralhd}, which predicts the 3D volumetric field of the hair, enabling a cohesive structure that captures both surface and depth information. By integrating this intermediate output with a 3D occupancy mesh based on the approach by Shen et al.~\cite{shen2020deepsketchhair}, we obtain a detailed, yet preliminary, hair representation. This comprehensive coarse structure provides a reliable foundation for the subsequent refinement stages in our framework.

\subsection{Synthetic Braid Modeling}
\label{sec:syn_braid}
We introduce a synthetic braid model trained to emulate patterns and the interwoven structure of a braid.
% We introduce a synthetic braid model and employ a learning strategy to ensure strand alignment and structural stability. The model is trained to emulate the characteristic patterns and interwoven structure of braided hair.

\subsubsection{Synthetic Braid Model}

% Based on the intertwined topologies of braids, a wide variety of complex braid hairstyles can be modeled by repeatedly applying rules derived from fundamental braid patterns.
% The typical structure of a braided hairstyle involves multiple strands, which can be further subdivided into smaller bunches. 
% These smaller bunches intertwine in specific configurations to form the overall braided pattern, enabling the modeling of diverse and intricate braid styles.

% Building on the systematical braid theory~\cite{Artin1950BraidTheory}, we adopt a set of generalizable rules to represent braid hairstyles based on the prior work~\cite{braid_syn_model}. 
% These rules involve adjusting the number of strands, modifying their crossover patterns, and merging additional strands at each crossing. 
% A braid can be effectively described by a series of mid-lines (or center-lines), denoted as $L_i$. 
% Consider a traditional three-strand braid (where $i=3$) as an example, where the structure of each braid bunch can be mathematically modeled using a combination of sinusoidal functions. The formulation for this standard braid pattern can be expressed as:

Based on systematic braid theory~\cite{Artin1950BraidTheory} and prior modeling work~\cite{braid_syn_model}, we adopt generalizable rules derived from fundamental braid patterns to represent diverse braided hairstyles. Complex braids typically consist of multiple hair strands subdivided into smaller bunches, intertwining through adjustable crossover patterns. We model each braid using mid-lines (center-lines) \( L_i \); for example, a traditional three-strand braid (\( i=3 \)) can be mathematically described by a combination of sinusoidal functions as follows:
\begin{equation}
\small
\begin{aligned}
L_0: & \ x = a \sin(t), \ y = t, \ z = b \sin(2t) \\
L_1: & \ x = a \sin(t + 2\pi/3), \ y = t, \ z = b \sin(2(t + 2\pi/3)) \\
L_2: & \ x = a \sin(t + 4\pi/3), \ y = t, \ z = b \sin(2(t + 4\pi/3)),
\end{aligned}
\label{eq:midlines}
\end{equation}
where $y$ represents the braiding direction, and constants $a$ and $b$ control the braid's amplitude and spacing. This model can adapt to various braid styles by adjusting the sine wave parameters.

% In this formulation, $y$ represents the braiding direction, while constants $a$ and $b$ control the amplitude and spacing of each braid bunch’s shape. This model is adaptable to various braid types by modifying sine wave parameters according to different strand configurations.

% Drawing inspiration from procedural techniques~\cite{xiao2021sketchhairsalon}, we leverage these braid-generating rules to create a learning-based braid model. In our approach, parameters $a$, $t$, $b$, and $r$ are defined as learnable, allowing for flexible adjustments that enhance realism of braid reconstruction. To add natural variation, we further incorporate a noise term in the formulation. In image coordinates, each braid bunch $L_i (i=0, 1, 2)$ in our synthetic three-strand model is defined as:

Inspired by procedural techniques~\cite{xiao2021sketchhairsalon}, we leverage braid-generating rules to build a synthetic braid model with learnable parameters ($a$, $t$, $b$, $r$). In our synthetic three-strand model, each braid bunch $L_i (i=0,1,2)$ is defined in image coordinates as:
\begin{equation}
\small
\begin{aligned}
x &= a \sin(wt + 2i + 2\pi/3) + x' \\
y &= y' \\
z &= b \sin(2 (w t + 2i \pi / 3)) + z' \\
r &= (1 + \text{noise}_i) \cdot \text{radius},
\end{aligned}
\label{eq:syn_eq}
\end{equation}
where $t$ is the frequency vector, $a$ defines braid width, and $b$, $w$, $\text{radius}$ are scalar constants. Adjustments $x'$, $y'$, $z'$ serve for spatial positioning and $\text{noise}_i$ introduce spatial variation for enhancing realism. %Our flexible parameterization generalizes easily to diverse braid patterns.

% where $t$ denotes the frequency vector, $a$ specifies the width vector corresponding to each $t$, and $b$, $w$, and $\text{radius}$ are scalar constants.
% The terms $x'$, $y'$, $z'$, and $\text{noise}_i$ serve as adjustments for spatial positioning and introduce slight variations, enhancing the realism of the synthetic strands. 
% This flexible, parameterized model is readily generalizable to various braided hairstyles, effectively capturing the essential structural characteristics of braids and enabling straightforward application to a wide range of braid patterns.

% -------------------------------------------
\subsubsection{Braid Model Training}
% 1. 2D braid mask(SAM(point)->hair; SAM(automask)->braid mask)
% \noindent \textbf{2D Braid Mask Acquisition.} 
% We employ the advanced Segment Anything Model (SAM)~\cite{SAM} to segment the hair region. To separate the hair from the body, we provide the SAM-Predictor with an initial point, typically positioned near the upper center of the image. After isolating the hair, additional points are added to refine the segmentation of the braid itself. This step is essential for accurately distinguishing the intertwined structure of the braid from other hair sections, ensuring precise capture of the braid's 2D position and shape.

% 2. braid mask->mid-line
% \noindent \textbf{Mid-Line Identification.}
% As shown in Eq.~\ref{eq:midlines}, an isolated braid can be represented by its mid-line, which follows the central path of the braid, capturing its curvature and varying width. To extract this mid-line from a 2D braid mask, we use the Zhang-Suen Thinning Algorithm\cite{thinning}, an iterative method that reduces binary images to skeletonized forms by progressively removing boundary pixels while preserving the shape’s topological structure. This algorithm produces a thin line that accurately represents the mid-point of the braid structure.

% 4. Change hair strands to image coordinate and use 2D mask projection to get all strands belonging to braids
\noindent \textbf{Training Data Acquisition.}  
We obtain 2D braid masks using the Segment Anything Model (SAM)~\cite{SAM}, guided by manually specified points to distinguish braid regions.
Mid-lines (Eq.~\ref{eq:midlines}) representing braid curvature and width are manually approximated from these masks.
We then project the 2D braid mask onto the initial 3D hair estimation, isolating braid strands from unbraided regions. This ensures that our synthetic model accurately captures only the relevant braid structure.
% \noindent \textbf{3D Braid Strands Identification.}
% We can identify the braided and unbraided part of 3D hair strands by projecting 2D hair mask to the initial hair strands estimation. The isolated braid strand point cloud can be further extracted by masking out strands belonging to the braided part of 3D hair. This step enables us to focus on the specific strands that make up the braid, ensuring that our synthetic braid model includes only the relevant hair strands and excludes extraneous hair.

% -------------------------------------------
\noindent \textbf{Loss Functions.}
% The objective of training the synthetic braid model is to construct a braid that aligns with the 2D braided topological structure and accurately represents the 3D position of braid strands. This can be achieved by minimizing the following loss terms:
Our training objective is to align the synthetic braid model with the real braid structure, both in 2D topology and 3D positioning. This can be achieved by minimizing the following loss terms:

% // pc loss
\(\bullet\) \textit{Point Cloud Distance Loss:}
% We use the Chamfer distance~\cite{Barrow1977ParametricCA} to measure the point cloud distance between the synthetic braid model and the estimated braid strands, quantifying the alignment between the synthetic braid and the estimated hair strands. The point cloud distance loss term $L_\text{pc}$ is defined as:
We measure alignment between synthetic braid ($S_1$) and estimated hair strands ($S_2$) using Chamfer distance~\cite{Barrow1977ParametricCA} :
% \vspace{-5pt}
\vspace{-4pt}
\begin{equation}
\small
\begin{aligned}[t]
L_\text{pc} & = \frac{1}{|S_1|} \sum_{p_1 \in S_1} \min_{p_2 \in S_2} \|p_1 - p_2\|_2 + \\
& \frac{1}{|S_2|} \sum_{p_2 \in S_2} \min_{p_1 \in S_1} \|p_2 - p_1\|_2.
\end{aligned}
\label{eq:l_pc}
% \vspace{-5pt}
\end{equation}
% where $S_1$ represents the set of points in the synthetic braid model, $S_2$ denotes the set of estimated braid strand points, and $p_i (i=1, 2)$ indicates the point in the set of points $S_i$.

% // project synthetic model to 2D image plane -> edge
\(\bullet\) \textit{2D Projection Loss:}
% To ensure the consistency between the 2D projected shape of the synthetic braid model and the real braid, we incorporate an edge-based loss. This loss enforces alignment in 2D edge distribution between the projected synthetic model and the Canny edge image derived from the input RGB image. Inspired by \cite{xiao2021sketchhairsalon}, we obtain the projected 2D edge image from the 3D synthetic braid model by following a subtraction approach. Specifically, we subtract a scaled version of the 2D projected shape image with radius $r$ from another with radius $r/1.4$, according to the synthetic braid parametrization~\ref{eq:syn_eq}. Since we use unlabeled RGB images for hair reconstruction, we represent the real braid topology by applying the Canny edge detector~\cite{Canny} to the original RGB image. To effectively measure the similarity between the synthetic and real braid 2D edge images, we compute the 2D projection loss as the average binary cross-entropy loss over all pixels in both images. The projection loss term $L_\text{proj}$ is then calculated as:
We enforce alignment between the synthetic braid's projected edges and real braid edges obtained via the Canny detector~\cite{Canny}. Following~\cite{xiao2021sketchhairsalon}, we generate synthetic braid edges by subtracting scaled projections (radius $r$ and $r/1.4$). The loss is computed using pixel-wise binary cross-entropy:
\vspace{-5pt}
\begin{equation}
\small
\begin{aligned}[t]
L_\text{proj} = - \frac{1}{HW} \sum_{i=1}^{H} \sum_{j=1}^{W} ( I_{i,j} \cdot \log({I'}_{i,j}) + \\
(1 - I_{i,j}) \cdot \log(1 - {I'}_{i,j}) )
\end{aligned}
\label{eq:l_proj}
\end{equation}
% where $I'$ is the 2D projected edge image from the synthetic model, $I$ is the edge image obtained from input RGB image, $H$, $W$ are the height and width of the images, and $i$, $j$ denotes the pixel position in the corresponding image.
where $I'$ and $I$ represent synthetic and real edge images, respectively, $H$, $W$ are height and width, and $i$, $j$ denote positions of the pixel.

% // regularization
\(\bullet\) \textit{Regularization Loss:}
% To ensure stability and prevent extreme variations in the synthetic braid model, we introduce regularization terms to keep synthetic braid model parameters within reasonable bounds. We particularly focus on controlling the parameter $b$ in the Eq.~\ref{eq:syn_eq}, as its drastic changes can affect the overall shape of the braid. Additionally, to promote a smooth transition of the braid strands along the z-axis (depth-wise), we include the standard deviation of the first-order derivative of the 1-D vector $z$. The regularization loss term $L_\text{reg}$ is formulated as follows:
To stabilize the synthetic braid model, we constrain parameter $b$ (Eq.~\ref{eq:syn_eq}) and enforce smooth depth transitions (along the z-axis) as:
% \vspace{-5pt}
\begin{equation}
\small
L_\text{reg} = \cdot \text{StdDev}(\frac{dz}{dt}) + \lambda |b - 10|
    \label{eq:l_reg}
\end{equation}
\begin{equation}
\small
\text{StdDev}(\frac{dz}{dt}) = \sqrt{\frac{1}{N-1} \sum_{i=1}^{N-1} \left( \frac{z_{i+1} - z_{i}}{\Delta t} - \overline{\frac{dz}{dt}} \right)^2}
    \label{eq:stddev}
\end{equation}
\begin{equation}
\small
\overline{\frac{dz}{dt}} = \frac{1}{N-1} \sum_{i=1}^{N-1} \frac{z_{i+1} - z_{i}}{\Delta t},
    \label{eq:avg_std}
\end{equation}
% \vspace{-5pt}
% where StdDev($\cdot$) is the standard deviation, $\frac{dz}{dt}$ is the first-order derivative of vector $z$, and $\lambda$ controls the smoothness constraint.
where $\frac{dz}{dt}$ is the first-order derivative of 1-D vector $z$, $\Delta t $ is the step size, $\overline{\frac{dz}{dt}} $ is the mean of the first-order derivative, and $\lambda$ controls the smoothness constraint.
% where $ z = [z_1, z_2, \dots, z_N] $ is the 1-D vector,$ \frac{dz}{dt} $ represents the discrete first-order derivative of $ z $,$ \Delta t $ is the step size between consecutive elements in $ z $, $ \overline{\frac{dz}{dt}} $ is the mean of the first-order derivative, and $\lambda$ is a hyper-parameter.

% -------------------------------------------
% Combining all components, the total loss function for training the synthetic braid model can be written as:
Combining all components, the total loss function can be written with weighting parameters $\lambda_1$, $\lambda_2$ as:
% \vspace{-5pt}
\begin{equation}
\small
L_\text{reg} = L_\text{pc} + \lambda_1 \cdot L_\text{proj} + \lambda_2 \cdot L_\text{reg},
    \label{eq:l_total}
% \vspace{-5pt}
\end{equation}
% where $\lambda_1$ and $\lambda_2$ are coefficients that control the weights of the edge loss and the regularization loss, respectively. This comprehensive loss function guides the optimization process, helping to achieve a well-aligned and stable synthetic braid model.

\noindent \textbf{Training Strategies.}  
Proper initialization is critical, as it strongly influences the final synthetic braid shape. Specifically, shift terms $x'$, $y'$, and $z'$ (Eq.~\ref{eq:syn_eq}) align the synthetic braid with estimated braid strands: $x'$ and $y'$ match mid-line pixel indices, while $z'$ aligns depth. We initialize $z'$ using the minimal depth from projected braid strands along the mid-line, facilitating faster convergence and accurate alignment. After optimization, we adjust the radius along the braid to ensure smoothness and prevent gaps or thickness inconsistencies.
% -------------------------------------------
\subsection{Real Braid Reconstruction}
\label{sec:real_braid}
We refine coarse strands guided by leveraging the optimized synthetic braid model.

\textbf{Full Braid Allocation.}
We first identify all relevant strands by measuring their minimum distance to the synthetic braid strands, using a radius slightly larger than the synthetic braid radius to ensure inclusion. Strands are allocated into braid bunches by balancing strand counts, assigning each strand based on proximity (using top points) to synthetic strands.

% To begin, we ensure that all relevant strands are included in the braid reconstruction process. This is achieved by calculating the minimum distance between each estimated braid strand and the synthetic braid strands. To capture all necessary strands, we set a radius slightly larger than that of the synthetic braid model to prevent omitting strands that belong to the braid structure.
% For strand allocation into specific braid bunches, we adhere to the principle that each braid bunch should have a balanced number of strands. Allocation is determined based on the distance to each synthetic strand and the balance of strand count in each bunch. We calculate the distance using only the top points of each strand, as lower segments are more prone to significant positional changes. Initially, strands within the set radius at the top of a synthetic strand are assigned to that strand. When a strand can be associated with multiple synthetic strands, it is allocated to maintain balance across the braid bunches.

% // reconstruction
\textbf{Braid Bunch Reconstruction.} 
We reconstruct braid strands iteratively from top to bottom, starting from the nearest points to its synthetic counterpart. Points deviating beyond the radius are adjusted to match the synthetic braid's orientation and radius constraints, proceeding downward until completion.
% We reconstruct the real 3D braid strands from top to bottom, guided by the synthetic braid model structure. For each strand, the starting point is the closest estimated point to the top of the corresponding synthetic strand. At each step, if a point falls outside the radius of its nearest point on the synthetic strand, it is adjusted to align in the same orientation as the synthetic strand, with a distance matching the radius. Subsequently, all points within the synthetic bunch (at distances no greater than the radius) are shifted by following the orientation vector between the current point and the next point along the synthetic strand. This iterative process continues until the newly reconstructed strand reaches the final two points of the synthetic strand.

% 2. attach the last part and replace the braid strands with reconstructed braid
\textbf{Braid Strands Replacement.}
After forming the primary braid structure, we attach the remaining segments from original strands to the reconstructed strands to ensure continuity. Original strands are then replaced, followed by down-sampling and smoothing to remove noise and achieve natural results.

% Once the primary braid structure is formed, we attach the remaining segment of the nearest existing strand to the new strand to ensure continuity throughout the reconstructed braid. The original braid strands are then replaced with the newly reconstructed strands. In addition, we apply post-processing techniques, including down-sampling and smoothing, to achieve a more natural appearance by reducing any noisy or wavy features in the reconstructed strands.

\section{Experiments}
\label{sec:exp}

% In this section, we evaluate the effectiveness of our methods through comprehensive ablation studies (Sec.~\ref{sec:ablation}), and compare our braided hair reconstruction results with other state-of-the-art methods (Sec.~\ref{sec:baselines}).

% We evaluate our method through comparisons with state-of-the-art approaches and ablation studies.

\textbf{Setup.}
Test images are sourced online and partially from SketchHairSalon~\cite{xiao2021sketchhairsalon}. As no ground-truth 3D data for real hairstyles and synthetic datasets lack braided hairstyles, we follow prior work~\cite{neuralhd} in using qualitative comparisons, enhanced with point cloud visualizations.
We train the synthetic braid model for 200 epochs using Adam~\cite{kingma2014adam} with an initial learning rate of $1\text{e}{-4}$, halved at epochs 100 and 133. We set $\lambda{=}1$ (Eq.\ref{eq:l_reg}), $\lambda_1{=}1\text{e}{-4}$, and $\lambda_2{=}1\text{e}{-3}$ (Eq.\ref{eq:l_total}). Parameters follow~\cite{xiao2021sketchhairsalon}: $w{=}1$, $t$ incremented by $0.05$ per point, $a{=}1/1.75$ of mid-line width, $b{=}10$, and radius $=7$. 

% The examples used in our experiments are sourced from a variety of online platforms and some of them are selected from SketchHairSalon dataset~\cite{xiao2021sketchhairsalon}.

% Regarding the training setup for our synthetic braid model, we set a learning rate of $1e{-4}$ and trained for a total of $200$ epochs using the Adam optimizer \cite{kingma2014adam}. The learning rate is reduced by half when the training epoch reaches $1/2$ and $2/3$ of the total epochs. For the hyper-parameter settings, in the Eq.~\ref{eq:l_reg}, $\lambda$ is set to $1$. In the total loss function~\ref{eq:l_total}, the 2D projection loss weight $\lambda_1$ is set to $1e{-4}$ and the regularization term weight $\lambda_2$ to $1e{-3}$.

% To ensure an effective initialization, we adopt parameters consistent with those employed in \cite{xiao2021sketchhairsalon}, facilitating an initial projection that closely resembles the target shape. Specifically, we initialize $w=1$ with $t$ starting at $0$, increasing by $0.05$ at each point. The number of samples for $t$ matches the number of mid-line points. The parameter $a$ is set to $1/1.75$ of the width between the 2D mid-line and its boundary, while $b$ is set to $10$ and radius to $7$. 

% training environment
% All experiments were conducted on a system with a single Nvidia 4070 GPU. 
% Training a synthetic braid model typically requires only a few minutes with input single-view RGB image and coarse hair estimation.

%%%%%%%%%%%%%%%%%%%
\begin{figure}[tbp]
    \centering
    \includegraphics[width=1\linewidth]{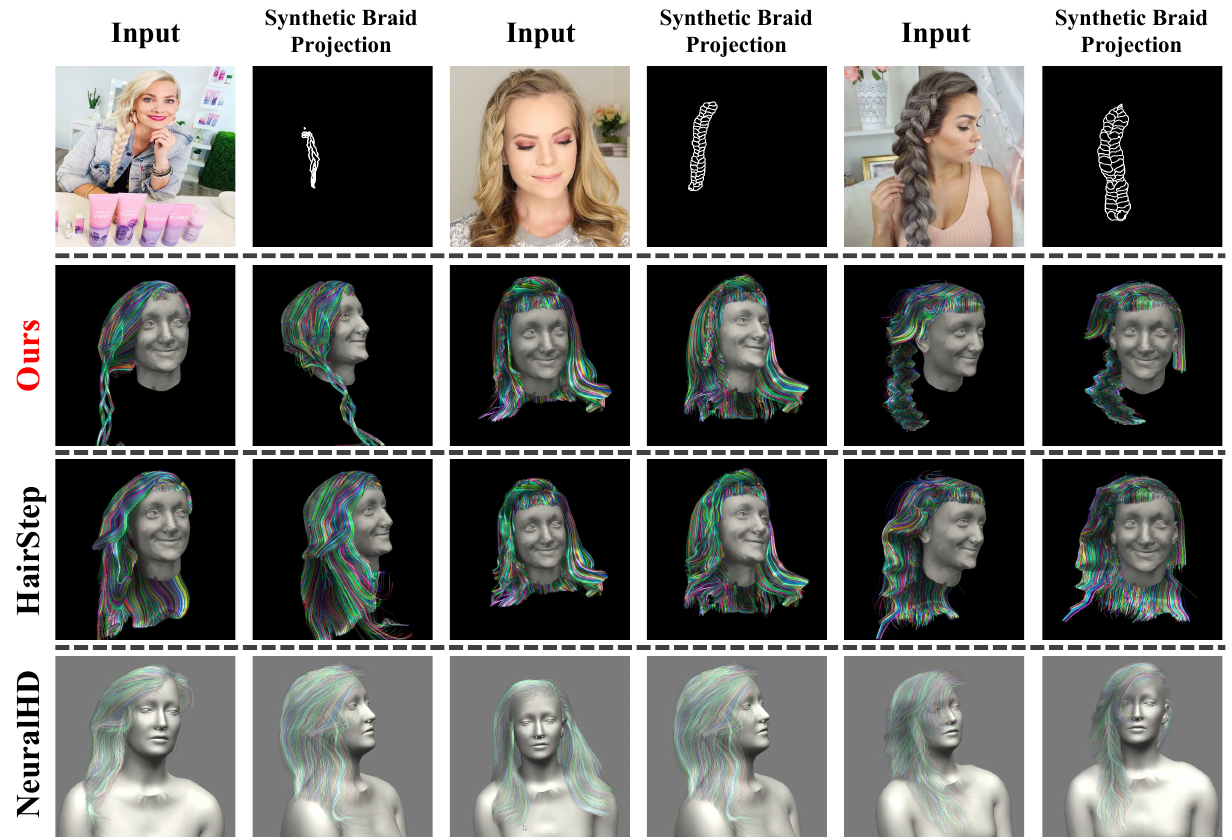}
    \small \caption{ Comparison of 3D braid reconstruction results among our method, HairStep, and NeuralHD on various input images with complex braided hairstyles. Each column shows the input image, the corresponding synthetic braid projection, and the reconstructions.}
    \label{fig:vis1}
    \vspace{-10pt}
\end{figure}

%%%%%%%%%%%%%%%%%%%%%%%
\subsection{Comparison to Baselines}

% We compare our method with state-of-the-art hair reconstruction approaches, namely HairStep~\cite{hairstep} and NeuralHD~\cite{neuralhd}, using single-view front-facing braided RGB images. As shown in Fig.~\ref{fig:vis1}, all methods are capable of capturing the overall hair shape and general hair orientation, particularly in regions with loose, unstructured hair. However, when addressing the intricate braid structure, existing methods fail to represent the intertwined patterns accurately. HairStep and NeuralHD primarily focus on estimating general hair volume and direction, which results in an inability to model the complex, interwoven geometry of braided hairstyles.

% In contrast, our method effectively approximates the 3D position, shape, and patterns of the braided hair region while maintaining alignment with the 2D projection of the input RGB image. This is achieved through our specialized strand-based modeling, which captures the key structural and geometric characteristics of braids.

% Our approach enables a more precise and realistic representation of braid structures by preserving the distinct woven patterns and their spatial arrangement, which are often overlooked or approximated poorly by other methods. This capability not only ensures improved fidelity in reconstructing braided hairstyles but also highlights the robustness of our method in handling challenging hair geometries.

We compare our approach against state-of-the-art hair reconstruction methods, HairStep~\cite{hairstep} and NeuralHD~\cite{neuralhd}, using single-view front-facing braided images. As illustrated in Fig.~\ref{fig:vis1}, existing methods successfully capture overall hair shapes but are notably struggling to accurately represent complex intertwined braid structures. In contrast, our method efficiently reconstructs detailed braid geometries directly from single-view RGB images without additional 3D inputs or training with labeled 3D braided hair data. Leveraging an efficiently trained synthetic braid model (typically within a few minutes), our strand-based representation effectively preserves intricate braid patterns and spatial alignment, achieving superior realism and robustness.

\begin{figure}[tbp]
    \centering
    \includegraphics[width=1\linewidth]{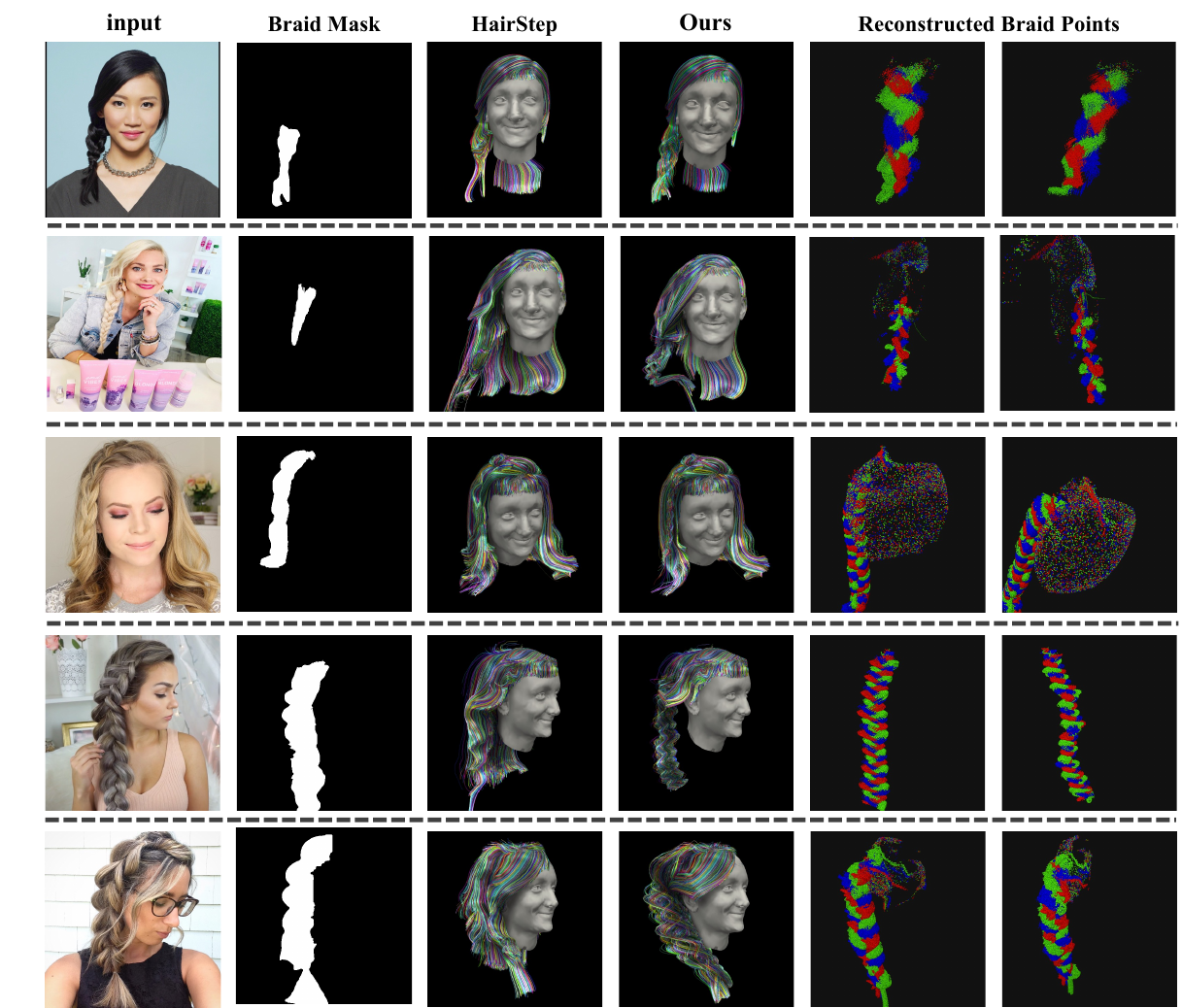}
    \small \caption{ Comparison of 3D braid reconstructed strands and Points on various input images with complex braided hairstyles. Each column shows input image, braid mask, reconstructed strands and points.}
    \label{fig:vis2}
    \vspace{-10pt}
\end{figure}

To further clarify and demonstrate the reconstructed braid structures, we present additional visualizations in Fig.~\ref{fig:vis2}. Due to the inherent limitations of rendering tools, certain intricate braid details may not be fully captured in standard renderings. Therefore, we provide colored point-cloud visualizations, where each braid bunch is distinguished using different colors (red, blue, and green). Two distinct viewpoints for reconstructed braid point clouds are shown: one matching the original RGB image perspective, and another from an alternate angle to clearly reveal the 3D intertwined geometry. These visualizations confirm our method’s capability to reconstruct structurally coherent and visually accurate braided hairstyles.
% To further demonstrate the reconstructed braid structure, we provide additional visualization results. Due to limitations of rendering tools, some intricate details of the intertwined braid structure may not be clearly visible in the main paper. Here, we include colored visualizations that highlight the reconstructed braid points, with each braid bunch represented by a different color.

% As shown in Figure~\ref{fig:vis2}, the braid point cloud consists of multiple strands, grouped into braid bunches corresponding to the synthetic braid model. Red, blue, and green colors are used to represent different braid bunches. Two views of the reconstructed braid points are presented: one from a viewpoint similar to the input RGB image and another from a different angle to better showcase the 3D intertwined structure. These visualizations demonstrate the capability of our method to reconstruct visually accurate and structurally coherent 3D braids.
\subsection{Ablation Study}
\label{sec:ablation}
% \subsubsection{Synthetic Braid Model Training}
\textbf{Losses Function.}
% // initialization
As shown in Eq.~\ref{eq:l_total}, our total loss combines point cloud distance, 2D projection, and regularization terms. Fig.~\ref{fig:syn+real_ablation} compares braid model alignment using different loss combinations. Using only point cloud and 2D projection losses causes boundary misalignment; replacing point cloud loss with regularization improves but does not fully resolve this issue. Combining point cloud and regularization losses achieves better strand alignment but still imperfect at the top. Integrating all three losses yields the best alignment, demonstrating that each loss—spatial consistency (point cloud), image alignment (2D projection), and structural stability (regularization)—is essential for accurate reconstruction.

\textbf{Post-processing Techniques.} 
Fig.~\ref{fig:downsample_smooth_ablation} compares 3D hair reconstructions with different post-processing strategies. Without post-processing, results exhibit noise and waviness. Applying smoothing first causes loss of key orientation details, resulting in 2D misalignment. Downsampling followed by smoothing preserves overall orientation, delivering a natural and visually coherent braid structure.

% Figure \ref{fig:downsample_smooth_ablation} illustrates the 3D hair model results with and without the application of downsampling and smoothing as post-processing techniques. Without any downsampling or smoothing, the braid reconstruction appears excessively noisy and wavy due to overemphasis on orientation changes inherent in the synthetic braid model. Conversely, applying smoothing before downsampling can result in the loss of critical orientation details, leading to mismatches between the 2D projected appearance and the input RGB image. From the comparison, it is evident that applying downsampling followed by smoothing effectively preserves the overall orientation and shape of the braid while achieving a natural and visually coherent pattern and appearance.

\section{Conclusion}

In this work, we present an efficient pipeline to reconstruct intricate 3D braided hairstyles from single-view images. Our method introduces a procedural synthetic braid model defined by sinusoidal functions and optimized through combined point cloud, 2D projection, and regularization losses. By refining coarse hair estimations using this model, our approach accurately captures complex intertwined braid structures and resolves occlusion-induced ambiguities. Experiments demonstrate superior accuracy and realism compared to state-of-the-art methods, highlighting our method's effectiveness for reconstructing complex hairstyles.

% In this paper, we addressed the challenge of reconstructing 3D braided hairstyles from single-view images, a task that is notably difficult due to the intricate interwoven structures and complex topologies of braids. To tackle these challenges, we proposed a novel pipeline that integrates a synthetic braid model and leverages a robust optimization strategy to achieve precise and realistic braid reconstructions.
% Our approach introduced a procedural synthetic braid model, defined by sinusoidal functions, which was optimized using a carefully designed loss function incorporating point cloud distance, 2D projection loss, and regularization. This model enabled accurate strand alignment and maintained a stable braid structure throughout the reconstruction process. By utilizing the trained synthetic model to refine coarse hair strands, our method effectively aligned the predicted braid structure while resolving ambiguities caused by occlusions. 
% Comprehensive experiments demonstrated that our method outperforms state-of-the-art approaches in reconstructing braided hairstyles, achieving superior accuracy and visual realism. This capability not only advances the representation of braided hair in digital human applications but also lays a foundation for future research in reconstructing complex hairstyles with intricate geometry and topology.

\clearpage

\bibliographystyle{plain}
\bibliography{main}

\end{document}